# TissUnet: Improved Extracranial Tissue and Cranium Segmentation for Children through Adulthood

**Author Names:** Markiian Mandzak[1,2,9*], Elvira Yang[1,2,8*], Anna Zapaishchykova[1,2,3*+], Yu-Hui Chen[5], Lucas Heilbroner[6], John Zielke[1,2,3], Divyanshu Tak[1,2,3], Reza Mojahed-Yazdi[1,2,3], Francesca Romana Mussa[1,2,3], Zezhong Ye[1,2], Sridhar Vajapeyam[1,4], Viviana Benitez[4], Ralph Salloum[15], Susan N. Chi[4,15], Houman Sotoudeh[14], Jakob Seidlitz[10,11,12,13], Sabine Mueller[7], Hugo J.W.L. Aerts[1,2,3], Tina Y. Poussaint[2,4], and Benjamin H. Kann[1,2+]

**Author Affiliations:**
1. Artificial Intelligence in Medicine (AIM) Program, Mass General Brigham, Harvard Medical School, Boston, MA, United States
2. Department of Radiation Oncology, Dana-Farber Cancer Institute and Brigham and Women's Hospital, Harvard Medical School, Boston, MA, United States
3. Radiology and Nuclear Medicine, CARIM & GROW, Maastricht University, Maastricht, the Netherlands
4. Boston Children's Hospital, Boston, MA, United States
5. Department of Data Science, Dana-Farber Cancer Institute, Boston, MA, United States
6. George Washington School of Medicine and Health Sciences, Washington, DC, USA
7. Department of Neurology, Neurosurgery and Pediatrics, University of California, San Francisco, United States
8. Ludwig Maximilian University of Munich, Munich, Germany
9. Ukrainian Catholic University, Lviv, Ukraine
10. Lifespan Brain Institute, The Children's Hospital of Philadelphia and Penn Medicine, Philadelphia, PA, 19104 USA
11. Department of Psychiatry, University of Pennsylvania, Philadelphia, PA, 19104 USA
12. Department of Child and Adolescent Psychiatry and Behavioral Science, The Children's Hospital of Philadelphia, Philadelphia, PA, 19104 USA
13. Institute for Translational Medicine and Therapeutics, University of Pennsylvania, Philadelphia, PA, 19104 USA
14. UT Southwestern Medical Center, Dallas, TX, USA
15. Department of Pediatric Oncology at Dana-Farber Cancer Institute, MA, USA

* First Co-authors
+ Correspondence

**Correspondence address to:**
benjamin_kann@dfci.harvard.edu
azapaishchykova@bwh.harvard.edu

**Abstract**
Extracranial tissues visible on brain magnetic resonance imaging (MRI) may hold significant value for characterizing health conditions and clinical decision-making, yet they are rarely quantified. Current tools have not been widely validated, particularly in settings of developing brains or underlying pathology. We present TissUnet, a deep learning model that segments skull bone, subcutaneous fat, and muscle from routine three-dimensional T1-weighted MRI, with or without contrast enhancement. The model was trained on 155 paired MRI–computed tomography (CT) scans and validated across nine datasets covering a wide age range and including individuals with brain tumors. In comparison to AI-CT-derived labels from 37 MRI–CT pairs, TissUnet achieved a median Dice coefficient of 0.79 [IQR: 0.77–0.81] in a healthy adult cohort. In a second validation using expert manual annotations, median Dice was 0.83 [IQR: 0.83–0.84] in healthy individuals and 0.81 [IQR: 0.78–0.83] in tumor cases, outperforming previous state-of-the-art method. Acceptability testing resulted in an 89% acceptance rate after adjudication by a tie-breaker(N=108 MRIs), and TissUnet demonstrated excellent performance in the blinded comparative review (N=45 MRIs), including both healthy and tumor cases in pediatric populations. TissUnet enables fast, accurate, and reproducible segmentation of extracranial tissues, supporting large-scale studies on craniofacial morphology, treatment effects, and cardiometabolic risk using standard brain T1w MRI.

**Key Results**
- TissUnet enables large-scale, automated analysis of extracranial tissues (skull, fat, and muscle) on T1w MRI with or without contrast.
- Validated on nine external datasets, including a blinded, randomized clinical evaluation study, with coverage of tumor and pediatric cases.
- Outperforms previous state-of-the-art (GRACE) with a median Dice of 0.83 (healthy) and 0.81 (tumor), compared to 0.73 and 0.60, respectively.

## 1. Introduction

Magnetic Resonance Imaging (MRI) is a widely used and standart imaging modality for visualizing brain anatomy, playing a central role in clinical care and neuroscience research. In particular, children and adults with neurologic conditions, such as brain tumors, multiple sclerosis, or dementia, undergo frequent MRI scans throughout diagnosis, treatment, and survivorship. In these instances, much focus is given to intracranial pathology, both qualitatively and quantitatively, motivating the development of many deep learning-based tools for intracranial brain and pathology segmentation (Stolte et al., 2024; Tierney et al., 2025).

Emerging evidence suggests that extracranial features may carry clinically meaningful, opportunistic information, including markers of treatment toxicity, physiologic reserve, and long-term outcomes(Cho et al., 2022; Hsieh et al., 2019; Zapaishchykova, Liu, et al., 2023; Zhang et al., 2023). Quantification and longitudinal tracking of these tissues would be clinically valuable, yet are impractical and challenging to do manually. Deep learning-based segmentation is a promising strategy for practical, accurate extracranial tissue segmentation, but there are no publicly available tools that enable comprehensive, three-dimensional segmentation of extracranial structures in standard brain MRIs, particularly for pediatric populations or those with brain pathology. This gap is especially relevant given the particular importance of sarcopenia in pediatric brain tumor (PBT) survivors, which leads to devastating physiologic frailty in up to 30% of survivors(Joffe et al., 2019) and is associated with reduced neurocognitive function, quality of life, and survival(Mager et al., 2023; Schulte et al., 2010). While physiologic frailty is well-characterized in adults, it remains poorly defined in children due to age- and puberty-related variability. Current clinical assessment relies on indirect metrics such as body mass index (BMI), which lacks specificity and correlates poorly with outcomes across pediatric populations (Marković-Jovanović et al., 2015).

Despite the growing interest in MRI-based body composition as a prognostic marker in these patients, few efforts have addressed the practical barriers to extracranial segmentation at scale. A major challenge is the ground-truth labels as manual annotation of extracranial structures is time-intensive and technically demanding (Galbusera & Cina, 2024). Although recent tools such as GRACE (Stolte et al., 2024) have demonstrated the potential of automated segmentation, they have not been validated in pediatric populations and in the presence of pathology. Further compounding these issues is the inconsistent application of defacing algorithms, which are commonly used in publicly shared MRI datasets to protect patient identity(Familiar et al., 2024). These methods vary widely in how much facial and extracranial tissue are removed, undermining model generalizability and limiting downstream biomarker discovery.

To address these challenges, we present a contrast-invariant T1w MRI deep learning (DL) framework, *TissUnet,* for automated segmentation of major extracranial tissues: bone(skull), subcutaneous fat, and muscle. In addition to volumetric analysis, our pipeline supports downstream anthropometric measurements of skull thickness derived from anatomical landmarks. We compare TissUnet against current state-of-the-art (SoTa) methods and multiple skull thickness estimation methods. To enable consistent analysis across heterogeneous, publicly available MRIs, we introduce a brain mask–guided region-of-interest (ROI) cropping strategy that isolates relevant extracranial structures while mitigating the effects of defacing and scanner variation. We further demonstrate its potential clinical application by modeling associations between tissue volume, body composition, and lipid profiles in adolescents.

## 2. Materials and Methods

TissUnet is a multitask, neural network trained to segment three extracranial tissues—skull, fat, and muscle **via 3-dimensional, T1-weighted (T1w) brain MRI (Figure 1A)**. Additionally, the model outputs 3D volumetrics of each tissue as well as an estimate of skull thickness(**Supplementary Methods 2).** The model was developed to perform accurately across the human lifespan and in the presence of intracranial pathology. To address variability introduced by defacing algorithms and scanner differences, the pipeline employs a novel brain mask–based region-of-interest (ROI) cropping pipeline(**Supplementary methods 6)**. To test the proposed method's robustness towards the registration deviations from the template, we rotate MRI T1w 5 degrees anterior and 5 degrees posterior and compare fat, muscle, and bone volumetric differences. Ten publicly available datasets were used for this study (**Figure 2; Supplement 1**) under data use agreements where necessary.

### *2.1 Training TissUnet*

TissUnet, based on the nnU-Net v2 framework (Wasserthal et al., 2023), was trained using a multi-center dataset SynthRAD2023, comprising 180 patients with brain tumors with co-registered CT-MRI T1w pre- and post-contrast imaging pairs (64% (N=115) male, mean age 65, range 3 - 93 years (Thummerer et al., 2023)(**Figure 2A).** Since muscle, fat, and bone are readily visible on CT, and due to the time and cost associated with generating de novo ground truth segmentations on MRI, we used segmentations generated by a previously validated CT-based AI algorithm, TotalSegmentator (Wasserthal et al., 2023), as initial ground truth labels that were then propagated to the co-registered T1w MRI (**Supplementary Methods 7).**

### *2.2 Evaluation Datasets*

Nine publicly available datasets were used for the evaluation (Figure 2; Supplementary Methods 1). The CERMEP dataset (Mérida et al., 2021), which consists of 37 adults of co-registered CT-MRI pairs (45.9% (N=17) male, mean age ± SD, 38.11 ± 11.36 years; range: 23–65 years) was used for evaluation of segmentation in healthy subjects, and the multi-center ACRIN dataset (64%(N=29) male, mean age ± SD: 57.2 ± 9 years, range: 29-77 years), which consists of subjects with newly diagnosed glioblastoma multiforme, was used to evaluate segmentation in the setting of brain pathology (tumors). Seven additional MRI datasets (Calgary(Reynolds et al., 2020), ICBM (Kötter et al., 2001), IXI (*IXI Dataset – Brain Development*, n.d.), ABCD (Casey et al., 2018a), PING (Rivkin et al., 2010) BabyConnectome (Howell et al., 2019), Brats-PEDS (Kazerooni et al., 2024)) were used to evaluate TissUnet in pediatric healthy and brain tumor settings(Supplementary Materials 1). Scans were co-registered to MRI age-dependent T1-weighted asymmetric brain atlases, generated from the NIH-funded MRI Study of Normal Brain Development (NIHPD, Fonov et al., 2011), using rigid registration and rescaled to 1-mm isotropic voxel size to preserve anatomical size differences (Lasso, 2017/2023).

### *2.3 Evaluation and Statistical Analysis*

All statistical analyses were done in R (v4.3.3). Between-group comparisons were conducted using the Mann–Whitney U test, with false discovery rate (FDR) correction for multiple comparisons. Categorical variables were compared using the Chi-Squared test. Two-sided p-values < 0.05 were considered statistically significant.

We evaluated TissUnet performance using four distinct experimental setups across nine external datasets (**Figure 2**). First, we compared TissUnet-predicted segmentations of the skull, fat, and muscle to reference segmentations generated from CT using TotalSegmentator and to the GRACE method(Stolte et al., 2024) on CERMEP dataset (N=37, **Figure 2B)**. All CERMEP AI-generated segmentations passed manual image QA. The performance was assessed using the Dice similarity coefficient and 95th percentile Hausdorff distance (HD95).

Second, we compared model outputs of TissUnet and GRACE(Stolte et al., 2024) to manual segmentations from an expert neuroradiologist (H.S., board-certified, 17 years of experience). We randomly selected 10 cases with paired MRI-CT imaging available, 5(50%) MRI T1w with a brain tumor (glioblastoma) from the ACRIN TCIA dataset("ACRIN-FMISO-BRAIN," n.d.), and 5(50%) T1w MRIs from the CERMEP dataset with no diagnosis (Mérida et al., 2021) **(Figure 2C).**

Third, we conducted an acceptability assessment in which two trained annotators (A.Z., L.H.) rated segmentation quality in blinded 3D review using a 5-point Likert scale, which was categorized into "Acceptable", "Unacceptable", and "Bad MRI" categories **(N=108, Supplementary Methods 5).** Inter-rater agreement was quantified using Gwet AC1 (Wongpakaran et al., 2013). Disagreements were resolved by a third reviewer (B.H.K., a board-certified radiation oncologist with nine years of experience**)**. Subjects were randomly selected and stratified by age, sex, and dataset origin: Calgary(Reynolds et al., 2020), ICBM (Kötter et al., 2001), and IXI(*IXI Dataset – Brain Development*, n.d.), to ensure the diversity in imaging protocols, scanner types, and developmental stages (N = 54, 50% female, median age = 23, IQR[5-27]). To assess model performance in a pediatric brain tumor scenario, we randomly selected N = 54 patients from the BRATS-Peds 2023, which contains multi-institutional MRI scans of children diagnosed with high-grade gliomas. (See **Figure 2D).**

Lastly, a reviewer (B.H.K.) compared TissUnet and GRACE segmentations in a blind review using Slicer 3D extension (SegmentationReview, (Zapaishchykova, Tak, et al., 2023). The reviewer was blinded both to the segmentation method and diagnostic status(N=45, Calgary(Reynolds et al., 2020), ICBM (Kötter et al., 2001), IXI(*IXI Dataset – Brain Development*, n.d.), ABCD(Casey et al., 2018a), PING(Rivkin et al., 2010) , BabyConnectome(Howell et al., 2019) Brats-PEDS(Kazerooni et al., 2024)), **Figure 2E).**

To assess associations between blood cholesterol levels and predictors—including BMI, extracranial tissue volumes, sex, and age—we used uni- and multivariable linear regression models. The Box-Cox transformation was applied to normalize the cholesterol distribution. Model assumptions, including linearity, normality of residuals, homoscedasticity, and absence of multicollinearity, were evaluated using diagnostic plots, the Shapiro–Wilk test, and variance inflation factors.

## *2.6 Application of TissUnet-Derived Volumetrics as Predictors of Cholesterol*

To demonstrate potential clinical application, we used the Adolescent Brain Cognitive Development (ABCD) Study (Casey et al., 2018b), a large-scale, multi-institutional cohort designed to investigate brain and health development across adolescents. We applied TissUnet-derived tissue volumes to model relationships between muscle, fat, BMI, and blood cholesterol levels in youth. We compared univariable and multivariable linear regression models. The outcome variable, total blood serum cholesterol, was Box-Cox transformed ($\lambda = 0.2$) to approximate normality and stabilize the variance. Predictors included Body Mass Index (BMI), volumetric measures of the muscle and subcutaneous fat (derived from TissUnet), sex, and age. Assumptions of linearity, normality of residuals, homoscedasticity, and multicollinearity were assessed through diagnostic plots and variance inflation factors.

## 3. Results
### 3.1. Segmentation Evaluation

We manually reviewed each sample in the SynthRad training dataset and removed 13.8% of T1-weighted MRIs (N = 25) due to imaging artifacts, including motion and blurring. In the AI-CT as ground-truth validation study, TissUnet median Dice in the external cohort of healthy adult subjects was 0.79 [IQR 0.77-0.81], compared to 0.5[0.48-0.54] for GRACE (p<0.001) with significant improvements in skull, fat, and muscle segmentation (**Figure 1B, Table 1**). In the second validation study using manual human expert annotations as ground truth, TissUnet's median Dice in the external cohort of healthy subjects was 0.83 [IQR: 0.83-0.84] and in the cohort with brain tumors was 0.81 [IQR: 0.78-0.83] (**Figure 1C, Table 2**), compared to 0.73[0.7-0.74] and 0.6[0.57-0.62] for GRACE, respectively. TissUnet showed consistently higher segmentation accuracy across different groups (see **Figure 3** for example segmentations).

In the acceptability testing, following adjudication by a tiebreaker for cases with disagreement, the final acceptability rates were 89% (N=289) "Acceptable," 10% (N=34) "Unacceptable," and 0.1% (N=1) "Bad Images" (**Figure 1D, Supplementary Methods 5**). Intra-rater agreement was higher in the healthy cohort compared to the brain tumor cohort across all tissue types (**Table S1**).

In the blinded review, TissUnet had 100% (N=45) of cases rated as acceptable, whereas GRACE had 16% rated as acceptable, with 84% (N=38) labeled as requiring revision or edits (Figure 1E).

### 3.2 Skull Thickness Methods Comparison

TissUNet yielded skull thickness measurements that more closely aligned with CT-based thickness than other MRI-based approaches both in the healthy and brain tumor patient cohorts at the default HU threshold (**Table 3**) and across various CT HU window settings (**Supplementary Methods 3**). TissUnet demonstrated excellent agreement across a range of skull thickness magnitudes, with a mean difference 0.31 mm in healthy and 0.57 mm in brain tumor cohort (**Figure 4**).

### 3.3 Rotation Ablation Study for Brain-ROI Cropping

When simulating minor registration errors by applying a ±5-degree tilt, estimated tissue volumes remained stable, with absolute average percentile changes of less than 3% across all classes and health groups (**Table 4**). Based on a sample size of 54 participants per group, the study had 94.6% power to detect a moderate effect size (Cohen's $d \approx 0.5$) between groups using a two-sided Wilcoxon signed-rank test (two-sided 0.05 type I error).

### 3.4. Application of TissUnet-Derived Volumetrics as Predictors of Cholesterol

In the ABCD study, 888 subjects had blood cholesterol and corresponding T1w MRI available. Median age was 11.9 years [IQR 11.3-12.4], 44%(N=389) female, median BMI 19.2[IQR 17.2-22.6], median blood cholesterol was 155 (mg/dL) [IQR 139-173]. 71% (N=631) of subjects had normal cholesterol (less than 170 mg/dL (*Hyperlipidemia in Children | Symptoms, Diagnosis & Treatment*, n.d.)). Median extracranial muscle volume was 39 cm$^3$ [IQR 33-47], and median extracranial subcutaneous fat volume was 110 cm$^3$ [IQR 79-172].

Temporalis muscle volume was significantly associated with blood cholesterol in the multivariable regression model, adjusted for age, sex, and subcutaneous fat (β = −4.23×10$^{−3}$, p = 0.014, **Figure 6**, **Supplementary Tables S2-S3** for uni- and multivariable models)**.**

**Discussion**

In this study, we present TissUnet, a deep learning-based model for automated segmentation of extracranial tissues—skull, subcutaneous fat, and muscle—from T1-weighted brain MRI. Compared to previously proposed methods (Stolte et al., 2024), which focused on broad tissue segmentation in adults, TissUnet was validated across both pediatric and brain tumor datasets. By leveraging pseudo-labels derived from the AI-CT-based segmentation method (Wasserthal et al., 2023), we mitigated the need for time-consuming, large-scale manual MRI annotations. TissUnet achieved high agreement with both AI-CT-based labels and human experts on five external datasets, demonstrating generalizability across different age groups and pathologies. The model achieved median Dice scores of 0.81 and 0.83 in healthy and brain tumor cohorts, respectively. We extend the TissUnet beyond segmentation and volumetric calculation of extracranial tissues, and propose an automated skull thickness estimation pipeline, broadening the utility for applications in cranial growth tracking and surgical planning.

While CT remains the clinical reference for skull thickness estimation due to its calibrated HU values, median skull HU decreases with age, from 800-850 HU in younger adults to 500-600 HU in older individuals (Delso et al., 2015; Schulte-Geers et al., 2011), our automated TissUnet-based skull thickness pipeline does not rely on specific HU thresholds and produces measurements comparable to those derived from CT across the lifespan. This approach enables accurate and efficient estimation on MRI, which is heavily utilized for tracking health conditions, such as cancer and neurologic diseases, due to its superior soft tissue contrast and absence of radiation exposure when compared to CT.

We found that TissUnet performed better across tissue segmentation tasks, particularly in pediatric and tumor cases, when directly compared to previous methods in both quantitative metrics and blinded clinical acceptability evaluation. In blinded review, all TissUnet outputs were rated acceptable, while 84% of segmentations from the previous state-of-the-art method required revision. We believe its superior performance stems from two key factors. First, the training data: compared to the previous method, which was trained exclusively on older adults from a single site, our model was trained on the multi-center SynthRAD2023 dataset, spanning a wider age range and greater anatomical variability. Second, the use of nnUNetV2, which incorporates advanced automated augmentation and is more robust to variations in MRI protocols, reducing the need for extensive image normalization. Notably, training on cases with pathologies did not impair performance on healthy brains. We hypothesize that many subtle pathologies resemble normal anatomy. In skull thickness estimation, traditional methods such as CHARM (Puonti et al., 2020), BrainSuite (Shattuck & Leahy, 2002), and SPM25 (Friston et al., 2006; Tierney et al., 2025) rely on geometric or probabilistic assumptions and skull surface meshes that often fail in the presence of abnormal anatomy. For instance, SPM125 uses voxel-wise statistical models, BrainSuite applies morphology-based techniques, and CHARM employs a mesh-based probabilistic atlas. FreeSurfer (FreeSurfer Developers, 2018), though widely used, does not explicitly segment the skull but instead creates a mask by extending the brain surface outward by 3 mm. In contrast, our model learns directly from imaging data, enabling it to adapt to anatomical variation and perform more accurately in real-world clinical settings.

To mitigate variability in measured volumes of extracranial tissues caused by defacing artifacts, we introduced a brain mask–based cropping strategy to define a consistent region of interest across subjects. MRI scan anonymization procedures (Familiar et al., 2024), which are commonly used to protect patient privacy, can inadvertently remove or alter extracranial tissues, making it difficult to measure consistently. The standardized ROI maintained spatial consistency across datasets and proved reliable for skull and muscle estimation. It is notable that given the relatively small volume of the subcutaneous fat in the extracranial region, small variations in registration alignment can translate into large changes in absolute percentage.

We demonstrated how TissUnet-derived extracranial tissue volume offers a biologically meaningful context in modeling lipid profiles during youth. Monitoring cholesterol during adolescence provides valuable insight for identifying early cardiometabolic risk (*Hyperlipidemia in Children | Symptoms, Diagnosis & Treatment*, n.d.), yet integrating imaging with biochemical markers at scale has been limited by manual segmentation constraints. In the ABCD study, 888 participants had both T1-weighted MRI and blood lipid data available, enabling population-level analysis of extracranial fat volumes—an approach previously infeasible without labor-intensive expert annotation. While the temporalis muscle has emerged as a validated T1w-based surrogate marker for sarcopenia, existing methods are largely restricted to 2D cross-sectional area (CSA) or manual estimation (Hsieh et al., 2019; Zapaishchykova, Liu, et al., 2023). However, beyond 2D temporalis muscle segmentation, no established pipelines currently exist for systematic volumetric analysis of extracranial tissues—including skull, muscle, and fat—despite their potential to yield additional prognostic insights. Such automated MRI-derived measures may be particularly valuable in pediatric populations who routinely undergo neuroimaging, including childhood cancer survivors and those with chronic neurologic conditions. Future studies should evaluate whether volumetric temporalis muscle measurements outperform CSA in clinical risk prediction and functional outcomes.

This study has several limitations. First, T1-weighted fat-saturated sequences, although beneficial for suppressing fat signals that may obscure intracranial structures, render extracranial fat largely invisible, making it challenging to validate fat segmentation on such scans. Consequently, the generalizability of our model to fat-saturated images remains uncertain. Second, while the model was designed to tolerate defacing, segmentation performance can degrade under extreme conditions. For instance, in the UK Biobank dataset, extensive cropping of the temporalis muscle restricts accurate volume estimation. Future work should explore domain adaptation strategies to extend model applicability to neurodegenerative conditions, such as Alzheimer's disease. Moreover, prospective clinical studies are needed to assess the downstream clinical relevance of automated extracranial tissue measurements.

**Conclusion**

We present TissUnet, a robust deep learning-based pipeline for segmenting extracranial structures—skull, muscle, and fat—in T1-weighted brain MRIs. Trained on pseudo-labels from CT data and validated across diverse pediatric and adult settings, including brain tumor datasets, TissUnet enables accurate tissue quantification and introduces automated skull thickness measurement. By addressing common challenges in extracranial segmentation, including defacing and limited annotation availability, our method supports comprehensive neuroimaging and anthropometric analyses for future applications in clinical research, growth assessment, and treatment planning.

## Data and Code Availability

The complete dataset (**Supplementary Material 1**) aggregated for this study contains primary datasets that differ widely in terms of their "openness," that is, their availability for secondary use without restrictions or special efforts by the team. Preliminary studies ranged from fully open and downloadable datasets in the public domain to more restricted datasets that could only be used for specific purposes, under separate agreements, or after special efforts had been made to provide data in shareable form. The model weights, training and testing code are available at https://github.com/AIM-KannLab/TissUNet.

## Author Contributions

Conceptualization and Study Design: M.M., E.Y., A.Z., B.H.K., Data collection/curation: M.M., E.Y., A.Z., J.L.Z., S.V., Investigation: M.M., E.Y., A.Z., B.H.K., L.H., Code, Software: M.M., E.Y., A.Z.; Methodology, Formal Analysis, Visualizations (Figures): M.M., E.Y., A.Z., Y.H.C., H.S., L.H., B.K. Data Interpretation: M.M., E.Y., A.Z., B.H.K.; Manuscript Writing—original draft: M.M., E.Y., A.Z. Manuscript Writing—review & editing: M.M., E.Y., A.Z., Y.H.C., L.H., J.L.Z., D.T., R.M.Y, F.R.M, Z.Y., S.V., V.B., R.S, S.N.C, J.S., S.M., A.N., B.H.K, T.Y.P., H.J.W.L.A; Project administration: A.Z., B.H.K., H.J.W.L.A.; Resources: B.H.K., H.J.W.L.A., T.Y.P., Supervision: A.Z, B.H.K.. All authors have substantively revised the work, reviewed the manuscript, approved the submitted version, and agreed to be personally accountable for their contributions.

## Role of the funding source

The funders had no role in study design, data collection, data analysis, data interpretation, or report writing.

## Declaration of Competing Interests

JS holds equity in and is a director of Centile Bioscience.

Table 1 Median Dice and HD95 [IQR] on 3 tissue classes and overall, comparing GRACE and TissUnet models vs. AI-CT annotations on N=37 cases from the CERMEP dataset (Mérida et al., 2021). IQR=interquartile range, HD95=The 95th percentile Hausdorff Distance.

|  | Dice, median [IQR] ↑ | | | | HD95, median [IQR] ↓ | | | |
| --- | --- | --- | --- | --- | --- | --- | --- | --- |
| **Model** | **Skull** | **Fat** | **Muscle** | **Overall** | **Skull** | **Fat** | **Muscle** | **Overall** |
| **TissUnet (ours)** | 0.87 [0.85, 0.88] | 0.65 [0.58, 0.68] | 0.86 [0.84, 0.88] | **0.79 [0.77, 0.81]** | 3.79 [3.53, 4.35] | 2.53 [2.22, 2.91] | 3.53 [3.08, 3.93] | **3.35 [3.16, 3.61]** |
| **GRACE** | 0.78 [0.77, 0.79] | 0.49 [0.44, 0.52] | 0.25 [0.21, 0.29] | **0.50 [0.48, 0.54]** | 3.61 [3.3, 3.95] | 2.24 [2.12, 2.34] | 3.06 [2.67, 3.41] | **3.02 [2.78, 3.25]** |

Table 2 Median, IQR Dice, and HD95 on 3 tissue classes, comparing GRACE and TissUnet models vs. Human Expert annotations on N=10 cases (N=5 brain tumor and N=5 healthy subjects). Highlighted in bold are the best results across tissue class and health status. IQR=interquartile range, HD95=The 95th percentile Hausdorff Distance.

|  |  | Dice, median [IQR] ↑ | | | | HD95, median [IQR] ↓ | | | |
| --- | --- | --- | --- | --- | --- | --- | --- | --- | --- |
| **Health Status** | **Model** | **Skull** | **Fat** | **Muscle** | **Overall** | **Skull** | **Fat** | **Muscle** | **Overall** |
| **Healthy** | TissUNet (ours) | 0.83 [0.83, 0.86] | 0.59 [0.58, 0.6] | 0.84 [0.84, 0.87] | **0.83 [0.83, 0.84]** | 1.41 [1, 1.73] | 3.46 [3, 3.74] | 1 [1, 1.41] | **1.73 [1.41, 1.73]** |
| | GRACE | 0.76 [0.75, 0.76] | 0.73 [0.7, 0.74] | 0.18 [0.16, 0.27] | **0.73 [0.7, 0.74]** | 5.1 [3, 5.48] | 6.71 [6.48, 7.14] | 87.07 [75.05, 89.99] | **6.71 [6.48, 7.14]** |
| **Brain Tumor** | TissUNet (ours) | 0.9 [0.89, 0.91] | 0.72 [0.7, 0.73] | 0.81 [0.78, 0.83] | **0.81 [0.78, 0.83]** | 1 [1, 1] | 5 [3, 19.52] | 1.73 [1.73, 2.24] | **1.73 [1.73, 2.24]** |
| | GRACE | 0.74 [0.71, 0.76] | 0.6 [0.57, 0.62] | 0.2 [0.18, 0.2] | **0.6 [0.57, 0.62]** | 3.74 [3, 5.38] | 20.4 [13.42, 28.05] | 85.15 [81.8, 88.92] | **20.4 [13.42, 28.05]** |

Table 3 Skull thickness mean thickness and absolute difference (in mm) in healthy cohort CERMEP (N=37), add brain tumor cohort ACRIN (N=5) for CT (reference, HU>471 (Delso et al., 2015)), TissUnet(our method), GRACE (Stolte et al., 2024), CHARM (Puonti et al., 2020), BrainSuite (Shattuck & Leahy, 2002), and SPM25 (Friston et al., 2006; Tierney et al., 2025), SD = standard deviation, HU = Hounsfield Units, CT= computer tomography.

|  | Healthy | | With brain tumor | |
|---|---|---|---|---|
| Tool | Mean Thickness (SD) | Absolute Difference (Tool – CT) | Mean Thickness (SD) | Absolute Difference (Tool – CT) |
| CT (Reference) | 5.60 (0.81) | - | 6.81 (1.17) | - |
| TissUNet (ours) | 5.46 (0.90) | 0.31 | 7.38 (1.22) | 0.57 |
| GRACE | 5.05 (0.71) | 0.60 | 5.61 (1.07) | 1.21 |
| CHARM | 4.48 (0.69) | 1.11 | 3.88 (0.84) | 2.93 |
| BrainSuite | 3.09 (0.79) | 2.51 | 5.74 (2.61) | 1.48 |
| SPM25 | 8.01 (1.48) | 2.52 | 9.93 (1.98) | 3.11 |

Table 4 Average volumetric differences (in cm² and as percentiles relative to the unmodified volume ("zero" tilt) mean ± SD) following simulated ±5° head rotations (forward and backward) across 108 MRI (Healthy: n = 54; Brain tumor: n = 54). Pairwise group differences were assessed using a two-sided Wilcoxon–Mann–Whitney U-test. SD=standard deviation.

| Health Status | Tilt angle | Skull | Fat | Muscle |
|---|---|---|---|---|
| Brain Tumor | +5 | 684.81 ± 563.88<br>0.23% ± 0.17%<br>(p=0.95) | 853.24 ± 736.78<br>0.78% ± 0.70%<br>(p=0.94) | 37.54 ± 50.70<br>0.13% ± 0.16%<br>(p=0.99) |
| Brain Tumor | -5 | 1493.69 ± 1258.68<br>0.50% ± 0.38%<br>(p=0.85) | 1128.59 ± 1352.88<br>1.04% ± 1.00%<br>(p=0.86) | 193.41 ± 178.75<br>0.70% ± 0.76%<br>(p=0.81) |
| Healthy | +5 | 1054.83 ± 933.30<br>0.28% ± 0.21%<br>(p=0.99) | 2453.22 ± 1454.75<br>2.10% ± 1.32%<br>(p=0.79) | 35.74 ± 48.98<br>0.09% ± 0.10%<br>(p=0.97) |
| Healthy | -5 | 2260.83 ± 1672.91<br>0.64% ± 0.44%<br>(p=0.78) | 3580.98 ± 2119.41<br>3.19% ± 2.06%<br>(p=0.74) | 311.24 ± 296.54<br>0.79% ± 0.64%<br>(p=0.79) |

# Figures

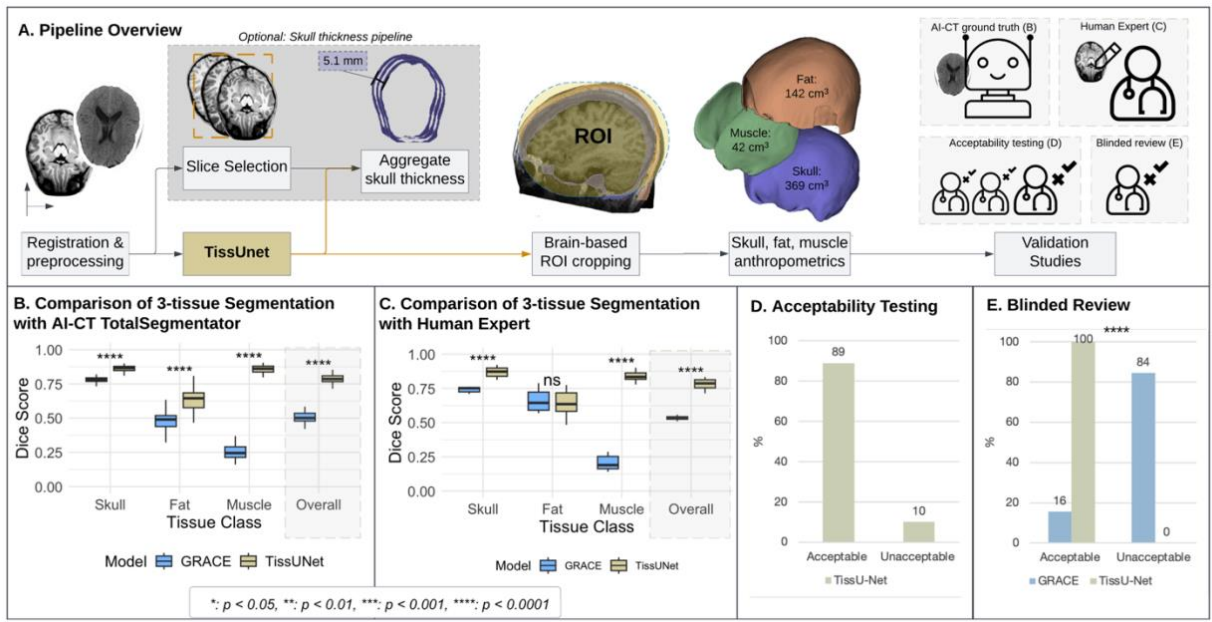

*Figure 1 A. TissUnet pipeline overview*. Step 1: The MRI T1w images are registered to the corresponding age-based NIHPD template. Step 2: TissUnet predicts 4 classes: brain, skull, subcutaneous fat, and muscle. Step 3: Using a brain mask, a universal ROI for calculating tissue volumes is created. Step 4: Skull, subcutaneous fat, and muscle anthropometric measurements are calculated on the cropped ROI. Optional: To estimate skull thickness, we used the DenseNet model to pick the top orbital roof slices and aggregate skull thickness measurements estimated from 95% measured tangents from 100 points at each 16x1 mm 2D axial slice. For validation studies, we created four different experiments: AI-CT ground truth(see **Figure 1B**), Human Expert annotations(see **Figure 1C**), Acceptability testing(**Figure 1D**) and Blinded Review(see **Figure 1E**).

*B. Boxplot of Dice of two DL methods performance compared to AI segmentations generated over CT images on a three-class tissue segmentation task*. TissUnet(yellow), GRACE(blue) for fat, bone, and muscle, and overall, between three classes on MRI T1w (N=37, healthy CERMEP dataset). See Table 1 for Dice and HD95 medians and IQR. Pairwise significance was tested with the Mann–Whitney U test with FDR correction for multiple testing. IQR= interquartile range

*C. Boxplot of Dice of two DL methods performance compared to human expert segmentations on a three-class tissue segmentation task*. TissUnet(yellow), GRACE(blue) for fat, bone, muscle, and overall between three classes on MRI T1w(N=10, 5 healthy and 5 brain tumor cases). The Dice score ranges from a minimum of 0 (worst score) to a maximum of 1 (best score). Box plots representing the interquartile range for each method per tissue. See Table 2 for Dice and HD95 by health status. Pairwise significance was tested with the Mann–Whitney U test with FDR correction for multiple testing.

*D. Bar plots for acceptability testing* (N=108 cases, 54 healthy and 54 brain tumor cases). See Table S3 and Figure S2 for intra-rater agreement scores by diagnosis and tissue class(skull, fat, muscle)

*E. Bar plot for blinded review* (N=54 MRI, 34 healthy and 12 with brain tumor). All cases (N=45, 100%) were rated acceptable using TissUnet, with none deemed unacceptable. In GRACE, 7 cases (16%) were rated acceptable and 38 cases (84%) were rated unacceptable. p-values measured using Chi-Squared.

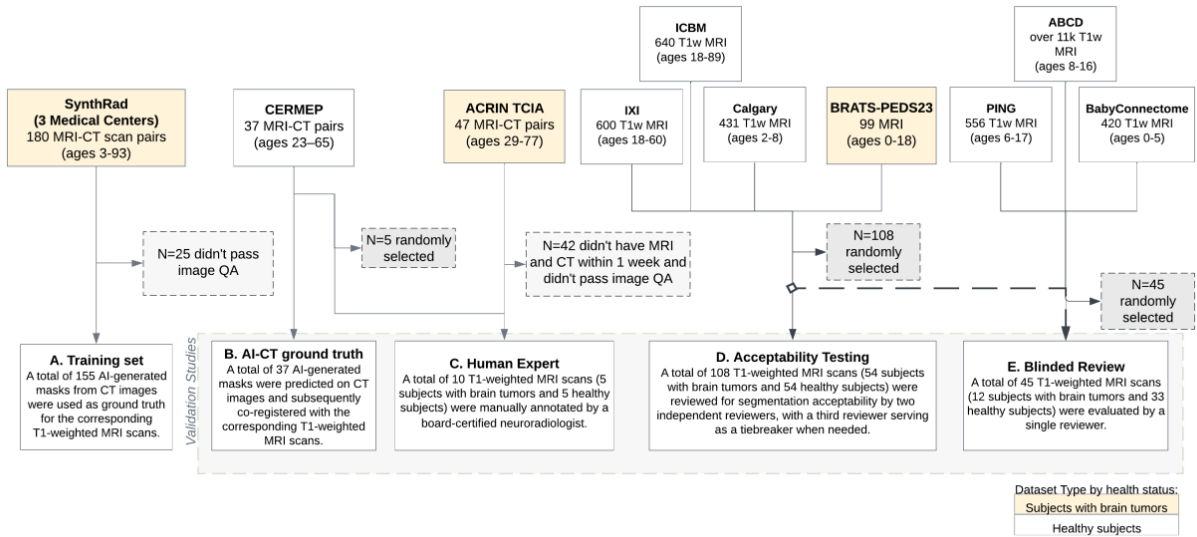

*Figure 2 Cohort selection by dataset origin and health status for TissUnet training and testing settings.* In yellow labeled datasets with subjects with brain tumors, in white- datasets with healthy subjects.

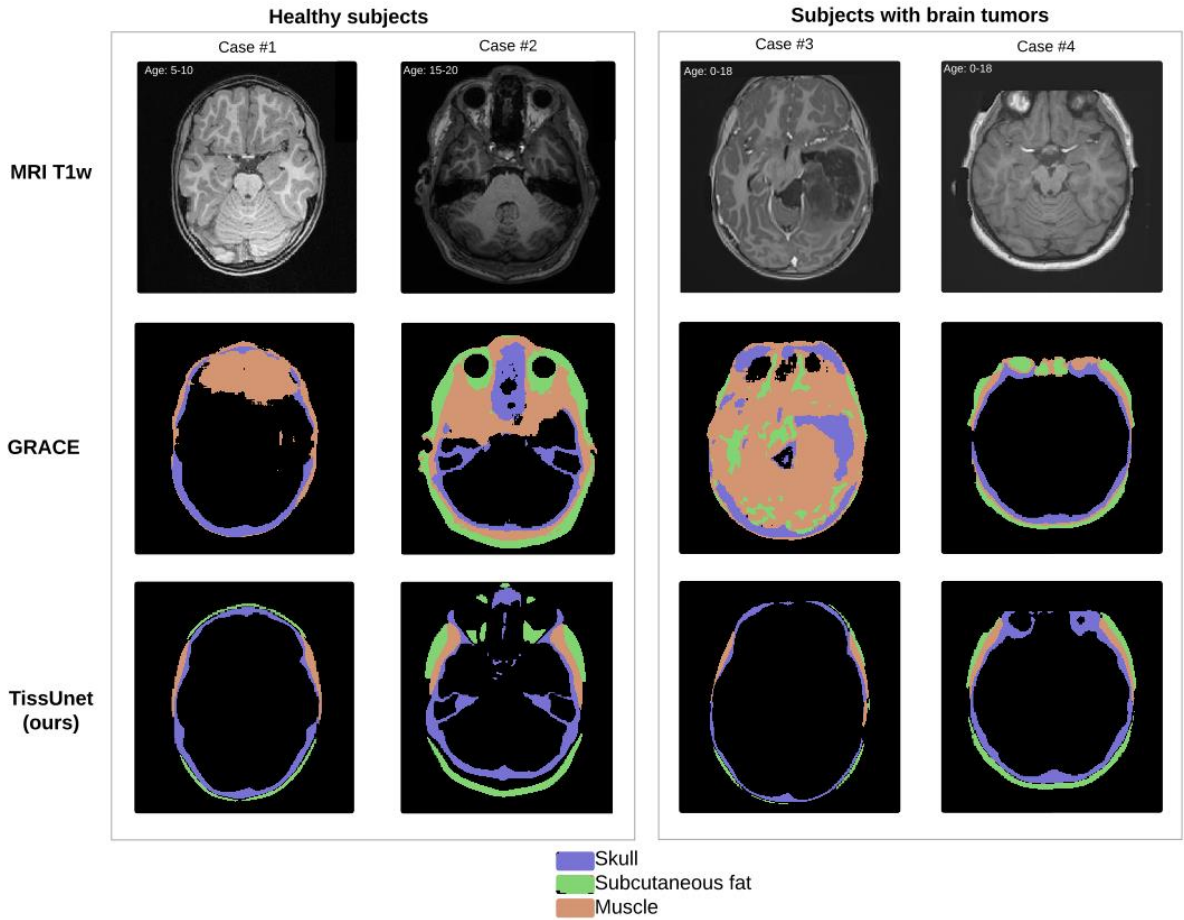

*Figure 3 Sample of four cases segmentations for three classes (skull in purple, muscle in green, and subcutaneous fat in orange) from the T1-MRI*. The results are shown in axial view. We compared GRACE(MRI T1w)(Stolte et al., 2024) to TissUnet(our) model in different age groups(overlaid in white text) and different health statuses: healthy(cases #1 and #2), and subjects with brain tumors(cases #3 and #4).

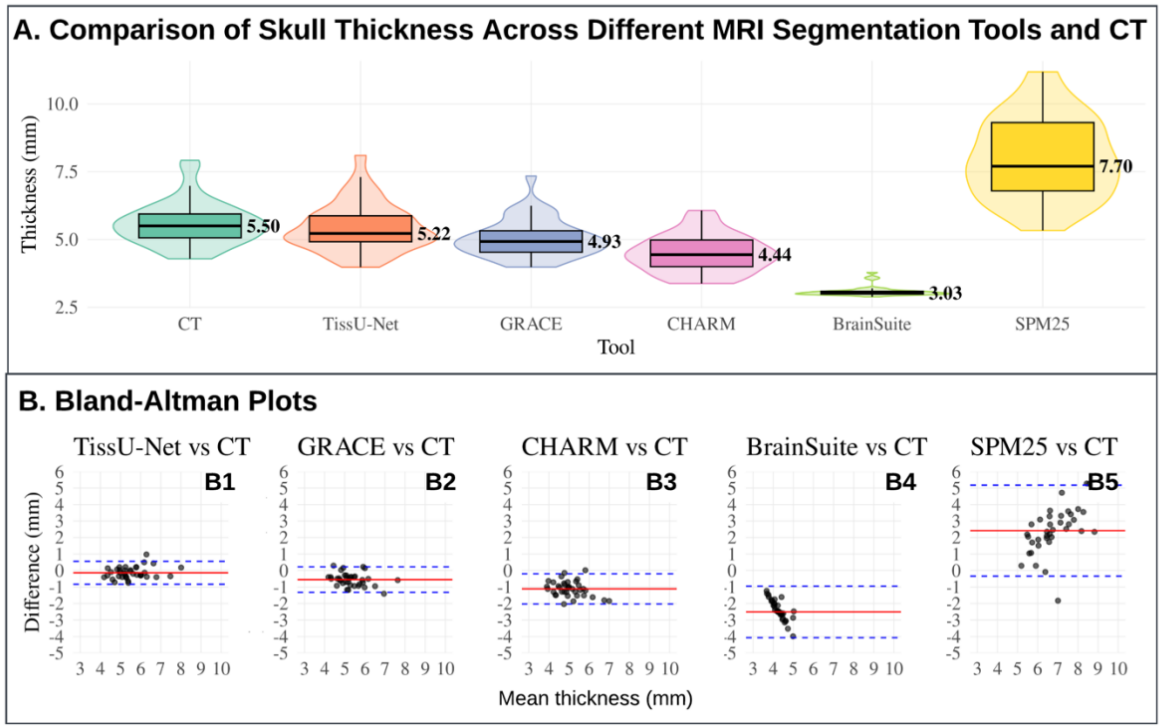

*Figure 4:* A. Comparison of median skull thickness between CT (reference, HU>471 (Delso et al., 2015)), TissUnet, GRACE (Stolte et al., 2024), CHARM (Puonti et al., 2020), BrainSuite (Shattuck & Leahy, 2002), and SPM25 (Friston et al., 2006; Tierney et al., 2025). The CERMEP dataset includes 37 healthy subjects with paired MRI T1-weighted (T1w) and CT images. The violin plots illustrate the overall distribution shape, with overlaid boxplots indicating the median (also labeled in text) and quartiles (See **Table 4**).

*B*. Bland–Altman plots compare each MRI-based segmentation tool (TissUnet, GRACE, CHARM, BrainSuite, SPM25) to the CT reference on the CERMEP dataset (Mérida et al., 2021). The solid red line represents the mean difference (bias), and the dashed blue lines indicate the 95% limits of agreement. CT = Computed Tomography, HU = Hounsfield units.

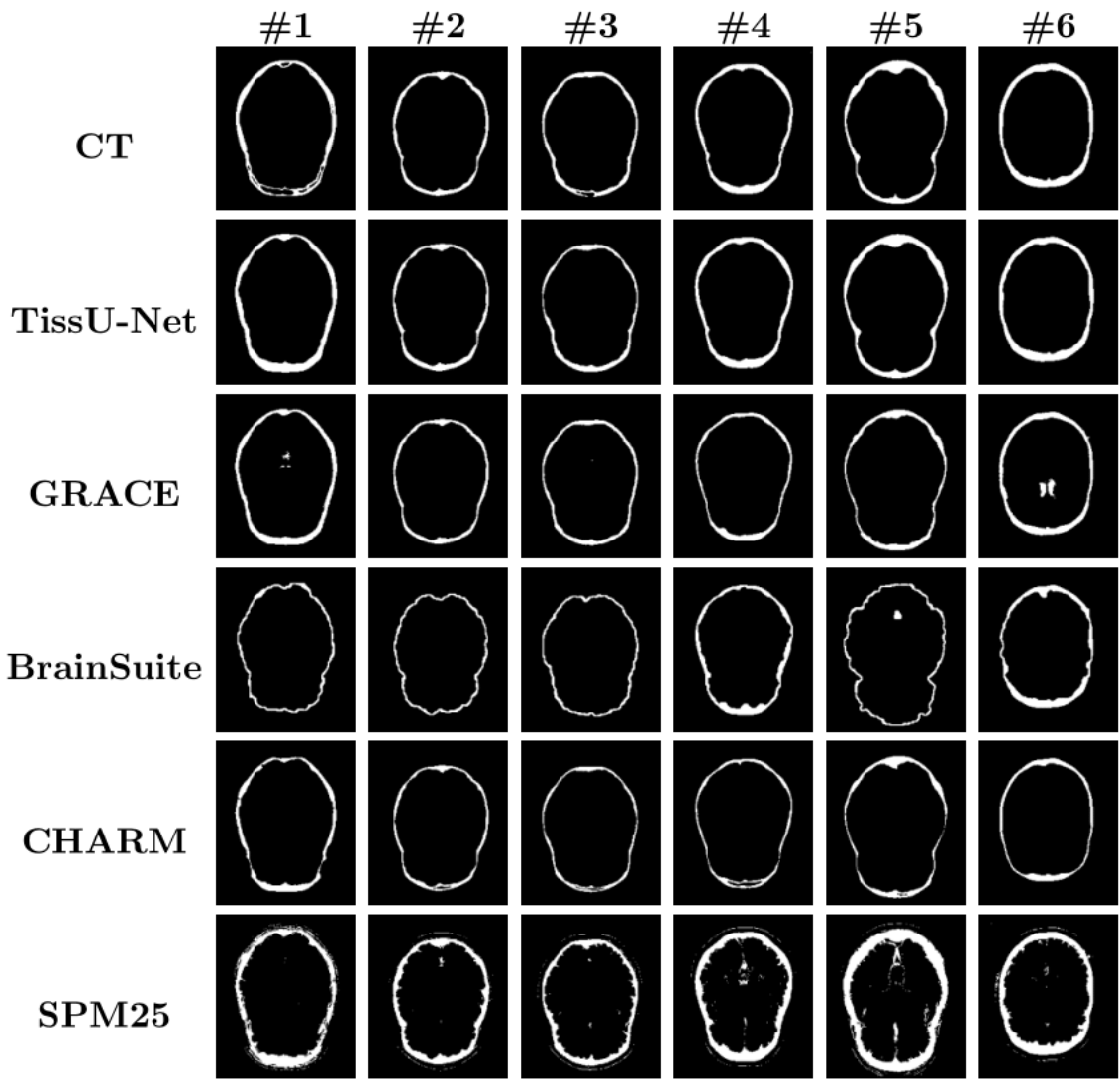

*Figure 5* Skull segmentations of three randomly selected healthy participants (columns #1-#3) from the CERMEP dataset and three randomly sampled patients with brain tumors (columns #4-#6) from ACRIN-TCIA (axial view) for CT ( HU>471 (Delso et al., 2015)) and T1w MRI-based segmentation models (TissUnet, GRACE (Stolte et al., 2024), CHARM (Puonti et al., 2020), BrainSuite (Shattuck & Leahy, 2002), SPM25 (Friston et al., 2006; Tierney et al., 2025))

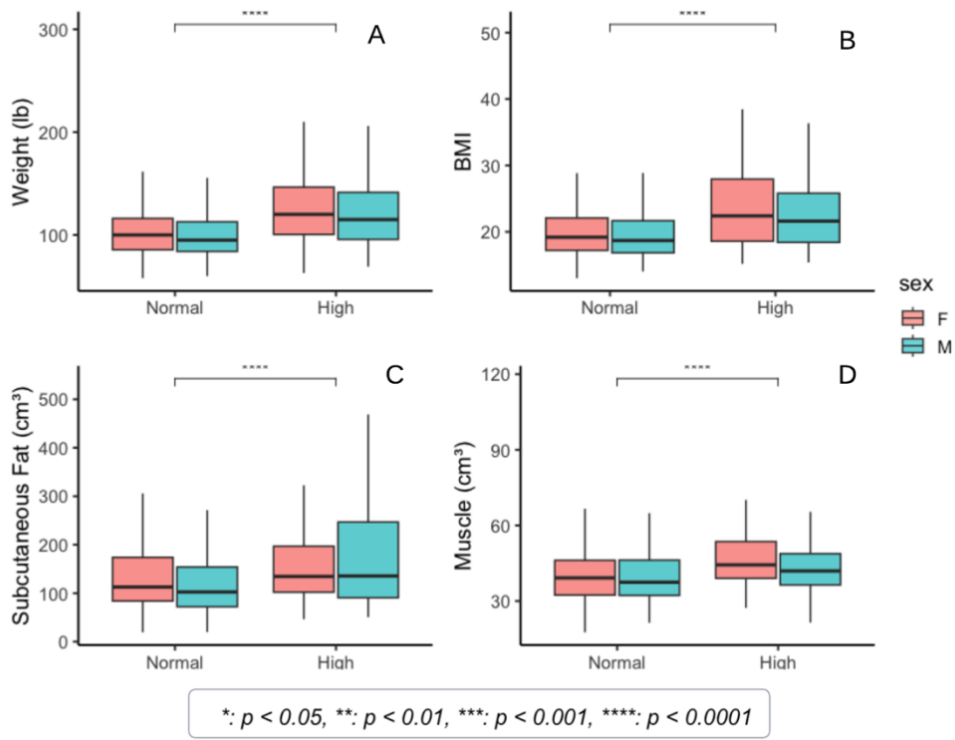

*Figure 6* **Associations between blood cholesterol levels and body composition metrics(BMI and weight) and extracranial volumetrics(muscle and subcutaneous fat), stratified by sex(N=888 subjects).** *Boxplots show distributions of weight (a), BMI (b), subcutaneous fat volume in cm$^3$ (c), and temporalis muscle volume in cm$^3$ (d) across binary cholesterol categories (below 170 mg/dl is "Normal" and above is "High"). Boxes represent the interquartile range (IQR), with the median shown as a horizontal line; whiskers extend to 1.5× IQR. Statistical comparisons between cholesterol groups were done using a two-sided Mann–Whitney U test with FDR adjustment for multiple comparisons. Asterisks indicate significance levels: p < 0.05(*), p < 0.01 (**), p < 0.001 (***); ns = not significant. IQR=interquartile range, BMI=body mass index.*

# Supplementary Materials
"TissUnet: Improved Extracranial and Cranium Tissue Segmentation for Children through Adulthood"

## Table of Contents

**Supplementary Methods 1. Datasets**
**CERMEP**
CERMEP is multi-modal database of 37 healthy subjects constructed with MRI, CT and [$^{18}$F]FDG PET images. For all participants, the PET/CT scan and the MRI session took place on the same day (between 8 a.m. and 14 p.m.). PET and CT data were acquired on a Siemens Biograph mCT64. The subjects' MR and PET images were visually reviewed by two neurologists for conspicuous brain abnormalities. MRI sequences were obtained on a Siemens Sonata 1.5 T scanner. Three-dimensional anatomical T1-weighted sequences (MPRAGE) were acquired in sagittal orientation (TR 2400 ms, TE 3.55 ms, inversion time 1000 ms, flip angle 8°). The images were reconstructed into a 160 × 192 × 192 matrix with voxel dimensions of 1.2 × 1.2 × 1.2 mm3 (axial field of view 230.4 mm). Sagittal Fluid-Attenuated Inversion Recovery (FLAIR, [15]) images (TR 6000 ms, TE 354 ms, Inversion time 2200 ms, flip angle 180°) were acquired with a 176 × 196 × 256 matrix and a voxel size of 1.2 × 1.2 × 1.2 mm3 (Mérida et al., 2021).

**ACRIN-TCIA**
Adult patients newly diagnosed with pathologically confirmed GBM (World Health Organization [WHO] grade IV) that had visible residual disease after surgical resection, and planned for initial treatment with radiation therapy (RT) and temozolomide (TMZ), with or without additional agents, were enrolled. Amount of residual tumor did not impact eligibility and visible residual disease included T2/FLAIR hyperintensity. The study enrolled the first patient in March 2010 and the last in August 2013, with follow up ending 1 year later (July 2014). Of the 50 patients enrolled, 42 had evaluable imaging MR studies and 38 patients had evaluable $^{18}$F-FMISO PET scans relating to the primary aim. Additionally, 37 patients had evaluable DSC imaging, 31 had evaluable DCE imaging, 39 had evaluable diffusion tensor imaging (DTI) data, 17 had evaluable spectroscopy (MRS) data and 13 patients had BOLD imaging that has never been analyzed. For each MR imaging session, patient scans were completed on 1.5 or 3 T scanners (Philips 3T (12 patients), GE 3T (12 patients), Siemens 3T (2 patients), and Siemens 1.5T (five patients) magnets). The current protocol can be found online ( [Protocol-ACRIN 6684 Amendment 7, 01.24.12](Protocol-ACRIN 6684 Amendment 7, 01.24.12)) ("ACRIN-FMISO-BRAIN," n.d.).

**BRATSPeds**
The BraTS-PEDs dataset includes a retrospective multi-institutional cohort of conventional/structural magnetic resonance imaging (MRI) sequences, including pre- and post-gadolinium T1-weighted (labeled as T1 and T1CE), T2-weighted (T2), and T2-weighted fluid attenuated inversion recovery (T2-FLAIR) im- ages, from 464 pediatric high-grade glioma. Inclusion criteria comprised of pediatric subjects with: (1) histologically- approved high-grade glioma, i.e., high-grade astrocytoma and diffuse midline glioma (DMG), including radiologically or histologically-proven diffuse intrinsic pontine glioma (DIPG); (2) availability of all four structural mpMRI sequences on treatment-naive imaging sessions. Exclusion criteria consisted of: (1) images assessed to be of low quality or with artifacts that would not allow for reliable tumor segmentation; and (2) infants younger than one month of age. Data for 464 patients was obtained through CBTN (n =120), DMG/ DIPG Registry (n = 256), Boston's Children Hospital (n = 61), and Yale University (n = 27) (Kazerooni et al., 2024)

**SynthRad**

This dataset consists of a total of 1080 CT and MRI/CBCT image pairs that were acquired between 2018 and 2022 in the radiation oncology departments of three Dutch university medical centers: University Medical Center Utrecht, University Medical Center Groningen, and Radboud University Medical Center. All patients in this dataset have been treated with external beam radiotherapy in the brain or pelvic region (photon or proton beam therapy). For anonymity, we will refer to the three centers with centers A, B, and C without specifying which letter belongs to which center. This dataset is presented as part of the SynthRAD2023 challenge (https://synthrad2023.grand-challenge.org/), which is structured into two tasks: task 1 addresses MR-to-CT image synthesis and hence consists of MR/CT image pairs, task 2 focuses on CBCT-to-CT image translation and consists of CBCT/CT image pairs. Two anatomical regions were considered for each task: the brain and the pelvis. Inclusion criteria were the treatment with radiotherapy and the acquisition of CT and either an MRI for treatment planning (task 1) or a CBCT for patient positioning during image-guided radiotherapy (task 2). Case selection in the brain was blind to clinical information concerning primary tumor etiology, making the tumor characteristics a random sample of the clinical routine. During data collection, no gender restrictions were considered, and the dataset consists of 64% male subjects and 36% female subjects. A mostly adult patient population was collected, with patients aged 3 to 93 years and a mean age of 65. For task 1, MRIs were acquired with a T1-weighted gradient echo or an inversion prepared—turbo field echo (TFE) sequence and collected along with the corresponding planning CTs for all subjects. The collected MRIs of centers B and C were acquired with a Gadolinium contrast agent, while the MRIs selected from center A were acquired without contrast (Table 2). No contrast was acquired for CT (Thummerer et al., 2023).

**ABCD**

Data used in the preparation of this article were obtained from the Adolescent Brain Cognitive Development SM (ABCD) Study (https://abcdstudy.org), held in the NIMH Data Archive (NDA). This is a multisite, longitudinal study designed to recruit more than 10,000 children age 9-10 and follow them over 10 years into early adulthood. The ABCD Study® is supported by the National Institutes of Health and additional federal partners under award numbers U01DA041048, U01DA050989, U01DA051016, U01DA041022, U01DA051018, U01DA051037, U01DA050987, U01DA041174, U01DA041106, U01DA041117, U01DA041028, U01DA041134, U01DA050988, U01DA051039, U01DA041156, U01DA041025, U01DA041120, U01DA051038, U01DA041148, U01DA041093, U01DA041089, U24DA041123, U24DA041147. A full list of supporters is available at abcdstudy.org [https://abcdstudy.org/federal-partners.html]. A listing of participating sites and a complete listing of the study investigators can be found at abcdstudy.org/consortium_members[https://abcdstudy.org/consortium_members/].

ABCD consortium investigators designed and implemented the study and/or provided data but did not necessarily participate in the analysis or writing of this report. This manuscript reflects the authors' views and may not reflect the opinions or views of the NIH or ABCD consortium investigators. The ABCD data repository grows and changes over time. The ABCD data used in this report came from the fast-track data release. The raw data are available at NDA [https://nda.nih.gov/edit_collection.html?id=2573]). Additional support for this work was made possible from supplements to U24DA041123 and U24DA041147, the National Science Foundation (NSF 2028680), and Children and Screens: Institute of Digital Media and Child Development Inc (Casey et al., 2018).

**PING**
This dataset was created by the NIH to support research on typical brain and behavioral development. It includes clinical, behavioral, and neuroimaging data collected from a large sample of children and adolescents between the ages of 4 and 18, recruited from multiple research sites across the United States. Participants aged 6 to 17 provided written assent to take part in the study. MRI scans were acquired using either General Electric or Siemens 1.5 Tesla scanners (Rivkin et al., 2010).

**Calgary**
The Preschool MRI study in The Developmental Neuroimaging Lab at the University of Calgary uses different magnetic resonance imaging (MRI) techniques to study brain structure and function in early childhood (OSF [https://osf.io/axz5r/files/osfstorage]). All imaging for this dataset was conducted using the same General Electric 3T MR750w system and 32-channel head coil (GE, Waukesha, WI) at the Alberta Children's Hospital in Calgary, Canada. Children were scanned either while awake and watching a movie, or while sleeping without sedation. The University of Calgary Conjoint Health Research Ethics Board (CHREB) approved this study (REB13-0020). T1-weighted images were acquired using an FSPGR BRAVO sequence with TR = 8.23 ms, TE = 3.76 ms, TI = 540 ms, flip angle=12 degrees, voxel size = 0.9x0.9x0.9 mm3, 210 slices, matrix size=512x512, field of view=23.0 cm. ASL images were acquired with the vendor supplied pseudo continuous 3D ASL sequence with TR = 4.56 s, TE = 10.7 ms, in-plane reso- lution of 3.5x3.5 mm2, post label delay of 1.5 s, and thirty 4.0 mm thick slices. The sequence scan time was 4.4 minutes (Reynolds et al., 2020)

**BabyConnectome**
The Baby Connectome Project (BCP [https://nda.nih.gov/edit_collection.html?id=2848]) is a four-year study of children from birth through five years of age, intended to provide a better understanding of how the brain develops from infancy through early childhood and the factors that contribute to healthy brain development. This project is a research initiative of the Neuroscience Blueprint – a cooperative effort among the 15 NIH Institutes, Centers, and Offices that support neuroscience research. The BCP is supported by Wyeth Nutrition through a donation to the FNIH. Images are acquired on 3T Siemens Prisma MRI scanners using a Siemens 32-channel head coil at the Center for Magnetic Resonance Research (CMRR) at the University of Minnesota and the Biomedical Research Imaging Center (BRIC) at the University of North Carolina at Chapel Hill (Howell et al., 2019)

**ICBM**
Data used in the preparation of this work were obtained from the International

Consortium for Brain Mapping (ICBM) database (ICBM [www.loni.usc.edu/ICBM]). The ICBM project (Principal Investigator John Mazziotta, M.D., University of California, Los Angeles) is supported by the National Institute of Biomedical Imaging and BioEngineering. ICBM is the result of efforts of co-investigators from UCLA, Montreal Neurologic Institute, University of Texas at San Antonio, and the Institute of Medicine, Juelich/Heinrich Heine University - Germany. Data collection and sharing for this project was provided by the International Consortium for Brain Mapping (ICBM; Principal Investigator: John Mazziotta, MD, PhD). The National provided ICBM funding Institute of Biomedical Imaging and BioEngineering. ICBM data are disseminated by the Laboratory of Neuro Imaging at the University of Southern California (Mazziotta et al., 2001)

**IXI**

The data [https://brain-development.org/ixi-dataset/] has been collected at three different hospitals in London: Hammersmith Hospital using a Philips 3T system (details of scanner parameters [http://brain-development.org/scanner-philips-medical-systems-intera-3t/]), Guy's Hospital using a Philips 1.5T system (details of scanner parameters [http://brain-development.org/scanner-philips-medical-systems-gyroscan-intera-1-5t/]), and Institute of Psychiatry using a GE 1.5T system (details of the scan parameters not available at the moment). The Thames Valley MREC granted ethical approval. The T1 and T2 images were acquired prior to diffusion-weighted imaging using 3D MPRAGE and dual-echo weighted imaging.(*The IXI Dataset*, n.d.)

**Supplementary Methods 2. Skull thickness estimation**
Skull thickness is typically calculated via CT, but is comparatively difficult to assess via MRI, given poorer bone-tissue contrast. For TissUnet-based skull thickness estimation, we extracted the contour from the TissUnet predicted bone mask using OpenCV Python (Figure 1A). We calculated median skull thickness by aggregating the central 95% of values (i.e, between the 2.5-97.5 percentiles) from over 100 tangents in each of sixteen consecutive 1mm 2D axial slices, beginning 10mm superior to the top orbital roof detected by the DL method(Zapaishchykova, Liu, et al., 2023) to offset frontal sinus cavities and orbital cavities. We compared the TissUnet median MRI skull thickness to CT-based thickness as a reference standard (default threshold set at 471 HU (Delso et al., 2015). We also compared performance across four other MRI-based methods for skull detection (GRACE (Stolte et al., 2024), CHARM (Puonti et al., 2020), BrainSuite (Shattuck & Leahy, 2002), and SPM25 (Friston et al., 2006; Tierney et al., 2025)).

## Supplementary Methods 3. CT-HU window comparison

Skull thickness was estimated from computed tomography (CT) using HU-based thresholding to segment cranial bone. To evaluate the sensitivity of skull thickness measurements to different segmentation thresholds, we compared multiple HU window settings: 300, 400, 500, and 800 HU on the CERMEP dataset (N = 37 healthy subjects, see **Figure S1**). These thresholds reflect the range of expected attenuation values for cancellous and cortical bone, which vary with age and bone mineral density(Delso et al., 2015; Schulte-Geers et al., 2011). Lower thresholds (e.g., 300 HU) are more inclusive of less mineralized or spongy bone typically seen in pediatric populations, whereas higher thresholds (e.g., 800 HU) emphasize dense cortical bone but may exclude thinner or under-ossified regions. One subject was excluded from the 800 HU due to anomalously high skull density and the absence of a valid segmentation contour at that threshold.

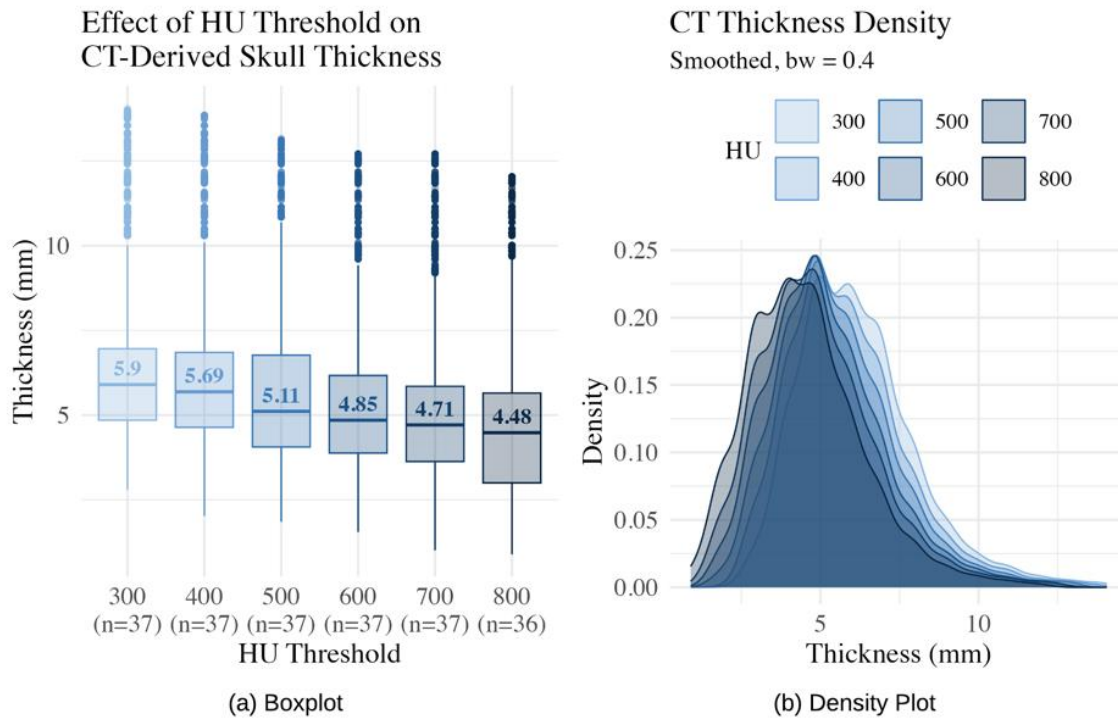

*Figure S2 Boxplot (a) and density plot (b) showing the distribution of median skull thickness measured from CT across different Hounsfield Unit (HU) threshold windows (300–800 HU) in the CERMEP dataset (N = 37 healthy subjects). One subject was excluded from the HU = 800 threshold due to a higher density skull and absence of a valid contour.*

**Supplementary Methods 4. Pseudo ground truth label generation**

To generate training pseudolabels for extracranial tissues, we applied the "head_muscles," "oculomotor_muscles," and "tissue_types" models from TotalSegmentator (Wasserthal et al., 2023) to CT scans from paired MRI–CT datasets. The raw labels were then merged into three target classes for training:
- **Muscle**: temporalis_left, temporalis_right
- **Bone**: skull
- **Subcutaneous fat**: subcutaneous_fat

We initially included in training an additional class, other muscle, which comprised smaller head and neck muscle groups(Walter et al., 2024). However, after applying a brain mask–guided region-of-interest (ROI) cropping strategy (see section **Materials and Methods *"2.4 ROI crop"***), this class frequently had zero visible volume. Due to its consistent exclusion from the cropped field-of-view, we opted not to evaluate or include this class in the final training or inference outputs.

## Supplementary Methods 5. Acceptability testing

Two reviewers evaluated each subject/tissue mask(total N=324 cases) using the following Likert score using SegmentationReviewer Slicer 3D extension (Zapaishchykova et al., 2023): (1) Acceptable with no changes, (2) Acceptable with minor changes, (3)Unacceptable with major changes, (4)Unacceptable and not visible and (5) Bad image. We then binned those into three groups: (1) Acceptable: Acceptable with minor changes + Acceptable with no changes, (2) Unacceptable: Unacceptable with major changes + Unacceptable and not visible, (3) Bad images. Each MRI scored differently by two raters, was reviewed by the Tie Breaker. Reviewer 1 classified 93% (N=300) of cases as "Acceptable" segmentations, 6% (N=20) as "Unacceptable," and 1% (N=4) as "Bad Images." Reviewer 2 labeled 69% (N=222) of cases as "Acceptable," 15% (N=49) as "Unacceptable," and 16% (N=53) as "Bad Images". Agreement between the Tie Breaker with Reviewer 1 was Gwet AC1 of 0.89 (95% CI: 0.86–0.93) and with Reviewer 2 Gwet AC1 of 0.72 (95% CI: 0.65–0.78) (See **Figure S2, Table S1**)

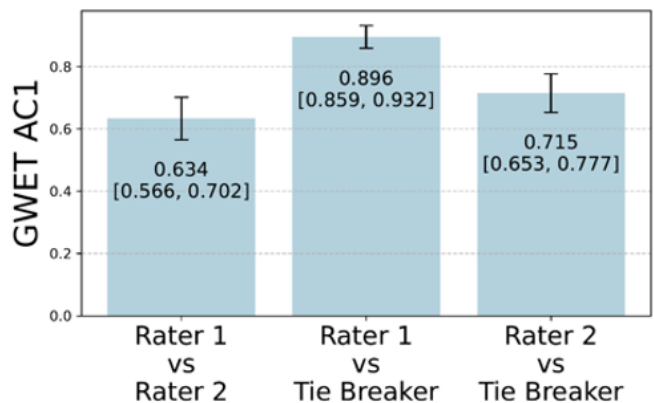

*Figure S2* ***Bar plots showing inter-rater agreement in acceptability testing, measured with GWET AC1*** with 95% CI confidence intervals bars for two raters(Rater 1 and Rater 2) and a Tie Breaker(N=108 cases, 54 healthy and 54 brain tumor cases). See Table 3 for GWET AC1 by diagnosis and tissue class(skull, fat, muscle)

Table *S*1. Inter-rater agreement in *acceptability testing on 3 tissue classes (skull, subcutaneous fat, and muscle) was measured using Gwet AC1(with 95 %CI in parentheses) within N=108 subjects (N=54 patients with brain tumor and N=54 healthy subjects). R1 = Rater 1, R2 = Rater 2, TB=Tie Breaker.*

| Health Status | Rater | Skull | Fat | Muscle | Overall |
|---|---|---|---|---|---|
| Brain Tumor | R1-R2 | 0.200 [-0.005, 0.405] | 0.699 [0.542, 0.856] | 0.751 [0.588, 0.913] | 0.573 [0.470, 0.676] |
| | R1-TB | 0.649 [0.443, 0.855] | 0.920 [0.838, 1.000] | 0.981 [0.943, 1.000] | 0.878 [0.822, 0.935] |
| | R2-TB | 0.457 [0.259, 0.654] | 0.791 [0.660, 0.921] | 0.777 [0.625, 0.929] | 0.687 [0.596, 0.778] |
| Healthy | R1-R2 | 0.371 [0.166, 0.576] | 0.798 [0.671, 0.925] | 0.852 [0.734, 0.971] | 0.693 [0.604, 0.783] |
| | R1-TB | 0.774 [0.635, 0.912] | 0.943 [0.876, 1.000] | 1.000 [1.000, 1.000] | 0.913 [0.865, 0.961] |
| | R2-TB | 0.462 [0.264, 0.660] | 0.852 [0.734, 0.971] | 0.852 [0.734, 0.971] | 0.742 [0.658, 0.826] |

**Supplementary Methods 6. Brain ROI**

To account for the defacing variability and different MRI scanners ROI between various open-source datasets, we introduce the brain mask-based region of interest (ROI) area cropping pipeline. It allows us to standardize the anatomical region definition for extracranial comparison measurements between subjects. For brain segmentation, we applied HD-BET (Isensee et al., 2019) to the MR images to generate brain masks. For each axial slice, we identify the first anterior pixel of the brain mask as a cut-off line, moving proximal and retaining everything below the cut-off line in the ROI.

Consider a three-dimensional Magnetic Resonance Imaging (MRI) volume denoted by:
$$V \in R^{H \times W \times D}, \qquad (1)$$

where *H, W,* and *D* correspond to the height (posterior-anterior axis), width (left-right axis), and depth (inferior-superior axis), respectively. Additionally, let $S \in \{0,1\}^{H \times W \times D}$ represent the binary brain segmentation mask associated with the volume *V*. In this mask:
$$S(x, y, z) = \begin{cases} 1, \text{if the voxel at coordinate } (x, y, z) \text{ is classified as brain tissue,} \\ 0, \text{otherwise} \end{cases} \qquad (2)$$

The binary mask *S* is obtained through a segmentation procedure and serves as a guide for the defacing process. The defacing procedure is executed on a slice-by-slice basis in the transverse (XY) plane. Processing starts at the inferior-most slice *(z=0)* and continues sequentially to the superior-most slice *(z=D−1)*. For each slice, we identify a dynamically adjusted cutting plane that delineates the most anterior part of the brain. All voxels positioned anterior to this plane are removed to eliminate any facial information. The cornerstone of the algorithm is the dynamic determination of the "top point," denoted as *T(z)* for each slice *z*. More formally, for a given slice *z*, the top point is defined as:
$$T(z) = \max\{x \in [0, H-1] \mid \exists\, y \in [0, W-1] \text{ such that } S(x, y, z) = 1\}, \qquad (3)$$

which represents the largest *x*-coordinate in that slice where brain tissue is present. In the event that a slice *z* does not contain any brain tissue (i.e., no voxel satisfies *S(x,y,z)=1*), we define:
$$T(z) = 0. \qquad (4)$$

To ensure consistent application of the defacing across slices, we introduce an adjusted top point *T\*(z)*, which propagates the most anterior brain boundary encountered up to the current slice. The recursive definition is:
$$T^*(z) = \max\{T(z), T^*(z-1)\}, \qquad (5)$$

with the initialization *T\*(−1)=0*. This recursive adjustment guarantees that the cutting plane does not regress even if subsequent slices exhibit a slightly more posterior brain boundary. With the adjusted top point *T\*(z)*, defined for each slice, the defacing procedure is executed voxel-wise. For each voxel at coordinate *(x,y,z)* in the segmentation mask S, the modified mask $\hat{S}$ is computed as:

$$\hat{S}(x, y, z) = \begin{cases} 0, \text{if } x > T^*(x), \\ S(x, y, z), \text{otherwise} \end{cases} \quad (6)$$

This rule removes any voxel that lies anterior to the defacing boundary (i.e., with an x-coordinate larger than $T^*(z)$) in the corresponding slice.

The same principle is applied to the original MRI volume $V$ to produce the defaced volume $\hat{V}$. Specifically, for each voxel $(x,y,z)$ in $V$, the defaced image is obtained as:

$$\hat{V}(x, y, z) = \begin{cases} 0, \text{if } x > T^*(x), \\ V(x, y, z), \text{otherwise} \end{cases} \quad (7)$$

By setting these voxels to zero, identifiable facial features are effectively removed while preserving the essential brain structures required for subsequent analysis (See **Figure S3**).

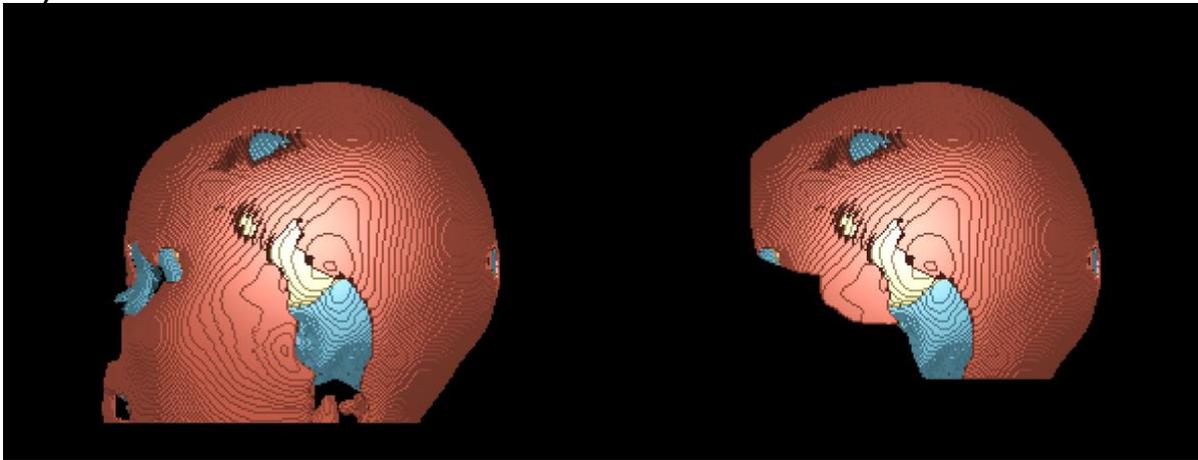

*Figure S3 Example of brain ROI cropping. The left panel shows a sagittal view of an original MRI scan and predictions, while the right panel displays the defaced counterparts.*

**Supplementary Methods 7. Model Training Specification**

TotalSegmentator is a model from the nnU-Net framework(Wasserthal et al., 2023), which is a UNet–based implementation that automatically configures all hyperparameters based on the dataset characteristics in a single-model setting. Based on the nnUNetV2 framework, the model was trained from scratch using Xavier initialization, Z-score normalization, a batch size of 2, and 128 × 128 × 128 patches. We used stochastic gradient descent with a learning rate of 1e-5 and a polynomial learning rate scheduler (exponent 0.9). The objective combined Dice and cross-entropy loss; performance was measured with pseudo-Dice. Augmentation followed nnU-Net's automated defaults. All MRI images and their corresponding segmentations underwent quality assurance (QA), and MRIs with artifacts were excluded from the training dataset. We trained nnUNetV2 optimizing the pseudo-dice and cross-entropy for 1000 epochs, batch size 2, on an NVIDIA GeForce RTX 4090, completing training in ~8.5 hours.

# Supplementary Tables

*Table S2 Univariable Linear Models for Box-Cox Transformed Cholesterol*

| Predictor | Estimate | 95% CI | p-value |
|---|---|---|---|
| BMI | 0.0026 | (-0.004, 0.0092) | 0.44 |
| Muscle (cm³) | -0.0039 | (-0.0067, -0.0012) | 0.0054 |
| Subcutaneous Fat (cm³) | -0.0001 | (-5e-04, 3e-04) | 0.54 |
| Sex, Male | -0.0496 | (-0.1123, 0.0131) | 0.12 |
| Age | -0.0053 | (-0.0093, -0.0013) | 0.0089 |

*Table S3 Multivariable Linear Model for Box-Cox Transformed Cholesterol*

| Predictor | Estimate | 95% CI | p-value |
|---|---|---|---|
| BMI | 0.0086 | (-0.0017, 0.0189) | 0.1 |
| Muscle (cm³) | -0.0042 | (-0.0076, -9e-04) | 0.014 |
| Subcutaneous Fat (cm³) | -0.0003 | (-8e-04, 2e-04) | 0.27 |
| Sex, Male | -0.0331 | (-0.0995, 0.0334) | 0.33 |
| Age | -0.0059 | (-0.0103, -0.0015) | 0.0081 |

*Table S4 Median and IQR by Sex and Cholesterol Group (below 170 mg/dl is "Normal" and above is "High"), N=888 subjects (Figure 6)*

| Sex | Cholesterol Level | | n | Median | IQR |
|---|---|---|---|---|---|
| F | Normal | BMI | 387 | 19.1900 | 4.88000 |
| F | High | BMI | 68 | 22.4000 | 9.35750 |
| M | Normal | BMI | 462 | 18.6850 | 4.82750 |
| M | High | BMI | 103 | 21.6100 | 7.39500 |
| F | Normal | Weight, lbs | 387 | 100.0000 | 30.47500 |
| F | High | Weight, lbs | 68 | 120.0500 | 46.00000 |
| M | Normal | Weight, lbs | 462 | 95.0000 | 28.77500 |
| M | High | Weight, lbs | 103 | 115.0000 | 45.65000 |
| F | Normal | Subcutaneous Fat (cm³) | 329 | 112.7250 | 90.02500 |
| F | High | Subcutaneous Fat (cm³) | 60 | 134.5030 | 94.83625 |
| M | Normal | Subcutaneous Fat (cm³) | 405 | 102.4660 | 81.94600 |
| M | High | Subcutaneous Fat (cm³) | 94 | 135.5530 | 156.07050 |
| F | Normal | Muscle (cm³) | 329 | 39.1980 | 13.77100 |
| F | High | Muscle (cm³) | 60 | 44.3705 | 14.55250 |
| M | Normal | Muscle (cm³) | 405 | 37.4760 | 13.97900 |
| M | High | Muscle (cm³) | 94 | 41.9515 | 12.40125 |

*PEDs) Challenge: Focus on Pediatrics (CBTN-CONNECT-DIPGR-ASNR-MICCAI BraTS-PEDs)* (arXiv:2404.15009). arXiv. https://doi.org/10.48550/arXiv.2404.15009

Mazziotta, J., Toga, A., Evans, A., Fox, P., Lancaster, J., Zilles, K., Woods, R., Paus, T., Simpson, G., Pike, B., Holmes, C., Collins, L., Thompson, P., MacDonald, D., Iacoboni, M., Schormann, T., Amunts, K., Palomero-Gallagher, N., Geyer, S., … Mazoyer, B. (2001). A probabilistic atlas and reference system for the human brain: International Consortium for Brain Mapping (ICBM). *Philosophical Transactions of the Royal Society of London. Series B*, *356*(1412), 1293–1322. https://doi.org/10.1098/rstb.2001.0915

Mérida, I., Jung, J., Bouvard, S., Le Bars, D., Lancelot, S., Lavenne, F., Bouillot, C., Redouté, J., Hammers, A., & Costes, N. (2021). CERMEP-IDB-MRXFDG: A database of 37 normal adult human brain [18F]FDG PET, T1 and FLAIR MRI, and CT images available for research. *EJNMMI Research*, *11*(1), 91. https://doi.org/10.1186/s13550-021-00830-6

Reynolds, J. E., Long, X., Paniukov, D., Bagshawe, M., & Lebel, C. (2020). Calgary Preschool magnetic resonance imaging (MRI) dataset. *Data in Brief*, *29*, 105224. https://doi.org/10.1016/j.dib.2020.105224

Rivkin, M. J., Ball, W. S., Wang, D.-J., McCracken, J. T., Brandt, M., Fletcher, J., McKinstry, R., Evans, A., Botteron, K., Pierpalo, C., & O'Neill, J. (2010). *Pediatric MRI* [Dataset]. NIMH Data Archive.

Schulte-Geers, C., Obert, M., Schilling, R. L., Harth, S., Traupe, H., Gizewski, E. R., & Verhoff, M. A. (2011). Age and gender-dependent bone density changes of the
43